\documentclass[11pt]{article}

\PassOptionsToPackage{table}{xcolor}

\usepackage[preprint]{acl}

\usepackage{times}
\usepackage{latexsym}
\usepackage[T1]{fontenc}
\usepackage[utf8]{inputenc}
\usepackage{microtype}
\usepackage{inconsolata}

\usepackage{amsmath}
\usepackage{amssymb}

\usepackage{graphicx}
\usepackage{tikz}
\usepackage{forest}

\usepackage{booktabs}
\usepackage{array}
\usepackage{multirow}

\title{Training-Free Inference Acceleration for Video Large Language Models: Methods, Trade-offs, and Evaluation
}

\author{
  Yuqi Li\textsuperscript{1,2,*},
  Chuanyue Yan\textsuperscript{3,*},
  Neil Rodrigues\textsuperscript{1},
  Xianglong Wang\textsuperscript{1} \\
  Xiaoqin Feng\textsuperscript{1},
  Junming Liu\textsuperscript{4},
  Xi Xiao\textsuperscript{5},
  Zhongwei Cheng\textsuperscript{1} \\
  Wei Li\textsuperscript{1},
  Lin Chen\textsuperscript{1},
  Yingli Tian\textsuperscript{2,\ensuremath{\dagger}} \\[0.5em]
  \textsuperscript{1}Wyze Labs, Inc.
  \textsuperscript{2}The City University of New York \\
  \textsuperscript{3}Wuhan Institute of Technology 
  \textsuperscript{4}University of Melbourne \\
  \textsuperscript{5}University of Alabama at Birmingham \\[0.3em]
  {\small
    \textsuperscript{*}Equal contribution.
    \quad
    \textsuperscript{\ensuremath{\dagger}}Corresponding author.
  }
}

\begin{document}
\maketitle
\begin{abstract}
Video Large Language Models (VLMs) have demonstrated strong capabilities in video understanding and reasoning, but processing long visual sequences incurs substantial computational, latency, and memory overhead. This survey examines training-free inference acceleration, focusing on reducing inference costs while preserving task-relevant spatiotemporal evidence. We organize existing methods by their intervention stages in the inference pipeline, covering input, visual encoding, projector, prefill, and decoding. Post-training quantization (PTQ) is discussed separately as a cross-stage extension with varying calibration requirements. We analyze how intervention position affects evidence preservation and computational savings, and why local efficiency improvements may not translate into end-to-end acceleration. We complement this analysis with quantitative evidence from prior studies and our own experiments under a shared evaluation setting to assess accuracy–latency trade-offs.
\end{abstract}
\begin{figure}[t]
    \centering
    \includegraphics[width=\columnwidth]{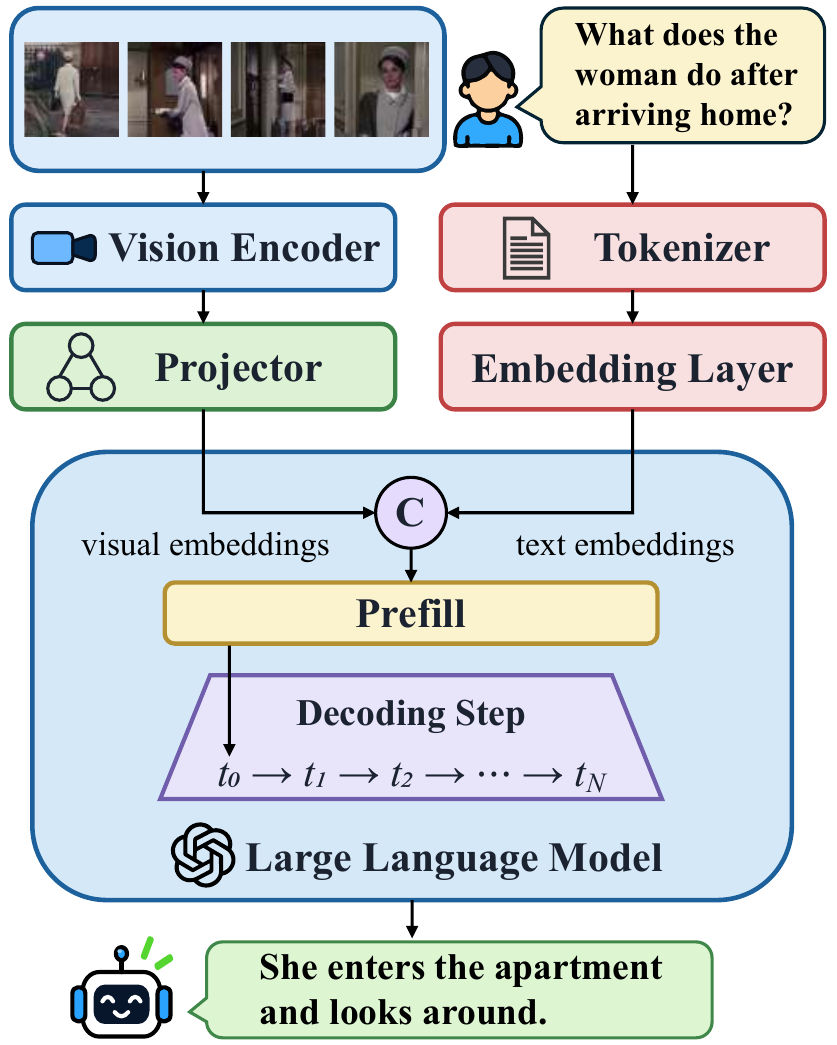}
    \caption{General architecture and inference pipeline of a VLM,
comprising a vision encoder, a projector connecting visual
and language representations, and an LLM with prefill
and decoding stages.}
    \label{fig:pipeline}
\end{figure}
\section{Introduction}

VLMs extend language-model reasoning to video by integrating
encoded visual representations with textual queries, as
illustrated in Fig.~\ref{fig:pipeline}.
Processing more frames or higher-resolution inputs increases
visual encoding and multimodal context processing costs
\cite{maaz2024video,zhang2025videollama,wang2025internvideo2,
bai2025qwen25vl,yuan2025tarsier2,li2025dytok,zhang2026efficient}.
However, video redundancy is non-uniform: brief actions may
provide critical evidence, while visually similar frames
may contain different task-relevant information
\cite{ye2025re,wang2025videotree,tang2025adaptive,zhang2025adard}.
Efficient VLM inference must therefore reduce redundant
computation while preserving the spatiotemporal evidence
required for reliable video understanding
\cite{wang2026earlytom,zhang2026efficient}.

Training-free acceleration seeks to reduce VLM inference
costs without training dedicated acceleration modules.
These methods operate across the inference pipeline:
input-stage methods reduce frames or spatial resolution
\cite{zhu2026focus,shi2026markey};
encoding-stage methods merge or reuse visual features
\cite{wang2026accelerating,wu2026metom};
projector- and prefill-stage methods reduce visual-token
or KV-cache overhead
\cite{ma2026apet,li2026segmo};
and decoding-stage methods reduce KV-cache access or
generation overhead
\cite{zeng2026hybridkv,kong2026parallelvlm}.
Intervention position affects both computational savings
and available information: earlier stages can avoid more
downstream computation, whereas later stages exploit
richer representations.
However, local savings do not necessarily translate into
end-to-end acceleration.

Existing surveys have examined multimodal efficiency from complementary
perspectives. Some provide broad reviews of efficient MLLM/LVLM architectures
and inference strategies
\cite{jin2025efficient,pathak2026towards,zhang2026efficient}, while others
focus on multimodal token compression or organize compression along the
inference lifecycle
\cite{shao2025survey,yao2026towards,wu2026data}. Recent work has also surveyed
training-free long-video understanding
\cite{liu2025towards}. 
However, the combined effects of intervention position
on spatiotemporal evidence preservation and realized
inference efficiency remain less explored in
training-free VLM acceleration.

To address these gaps, we examine where training-free
acceleration intervenes, how its benefits and limitations
vary across stages, and whether local savings translate
into end-to-end gains. Our contributions are threefold:
(1) we present a stage-wise taxonomy of training-free VLM
acceleration, covering input, visual encoding, projector,
prefill, and decoding, with PTQ discussed separately as
a cross-stage extension;
(2) we analyze how intervention position affects the
preservation of task-relevant spatiotemporal evidence
and potential computational savings; and
(3) we synthesize evidence from prior studies and conduct
experiments under a shared backbone and input protocol,
introducing Quality--Latency Balance (QLB) as a
baseline-normalized indicator of accuracy--latency
performance.

\section{Preliminaries}

\subsection{VLMs for Video Understanding}
\label{sec:pipeline}

Video understanding requires reasoning over temporally
distributed evidence, including object states, actions,
event order, and long-range dependencies. Unlike static
images, relevant evidence may span multiple frames and
depend on their temporal relationships
\cite{meng2026watch,an2026video}. VLMs encode such
spatiotemporal information for language-model reasoning
\cite{nguyen2024video}.

As illustrated in Fig.~\ref{fig:pipeline}, a typical VLM
consists of a vision encoder, a projector, and a large
language model (LLM). Given a video $\mathcal{V}$ and
query $\mathcal{T}$, input processing produces
$\mathcal{V}'=\mathcal{S}(\mathcal{V})$, followed by
$\mathbf{X}_v=E_{\phi}(\mathcal{V}')$ and
$\mathbf{H}_v=P_{\theta}(\mathbf{X}_v)
\in\mathbb{R}^{N_v\times d}$.
Let $\mathbf{H}_t$ denote the embedded query tokens.
In this simplified pipeline, the LLM receives
$\mathbf{C}=[\mathbf{H}_v;\mathbf{H}_t]$.
During prefill, the LLM processes this context,
constructs the KV cache, and produces the first output
token. Autoregressive decoding then accesses and updates
the cache to generate subsequent tokens.
Importantly, token compression after visual encoding
can reduce LLM computation but cannot recover the
encoding cost already incurred.

\subsection{Why Training-Free}
\label{sec:training_free}

Rapidly evolving VLM architectures make it costly to
retrain and maintain backbone-specific acceleration
modules \cite{wang2024qwen2vl,bai2025qwen25vl,
bai2025qwen3vl}. Training-free methods avoid such
retraining and can be more readily applied to existing
models \cite{yang2025topv,zhang2025vispruner}.
In this survey, training-free acceleration\footnote{
See Appendix~\ref{app:training_free_scope} for a
detailed discussion of its scope and boundaries.}
refers to methods that require no additional training
or fine-tuning to obtain their acceleration capability.
Frozen auxiliary models are permitted, but their
runtime overhead must be accounted for.
Methods requiring acceleration-specific training
are excluded from our core scope.

\definecolor{TaxGreen}{RGB}{92,142,66}    
\definecolor{TaxTeal}{RGB}{35,124,146}    
\definecolor{TaxBlue}{RGB}{47,111,220}    


\newcommand{\paper}[2]{#1~\citep{#2}}
\newcommand{\taxsup}[1]{\raisebox{0.36em}[0pt][0pt]{\fontsize{2.6}{2.6}\selectfont #1}}

\newcommand{\taxfont}{\fontfamily{ptm}\fontsize{6.0}{4.0}\selectfont}

\begin{figure*}[!t]
\centering
\begin{tikzpicture}
\node[anchor=south west, inner sep=0] (image) at (0,0)
  {\includegraphics[width=\textwidth]{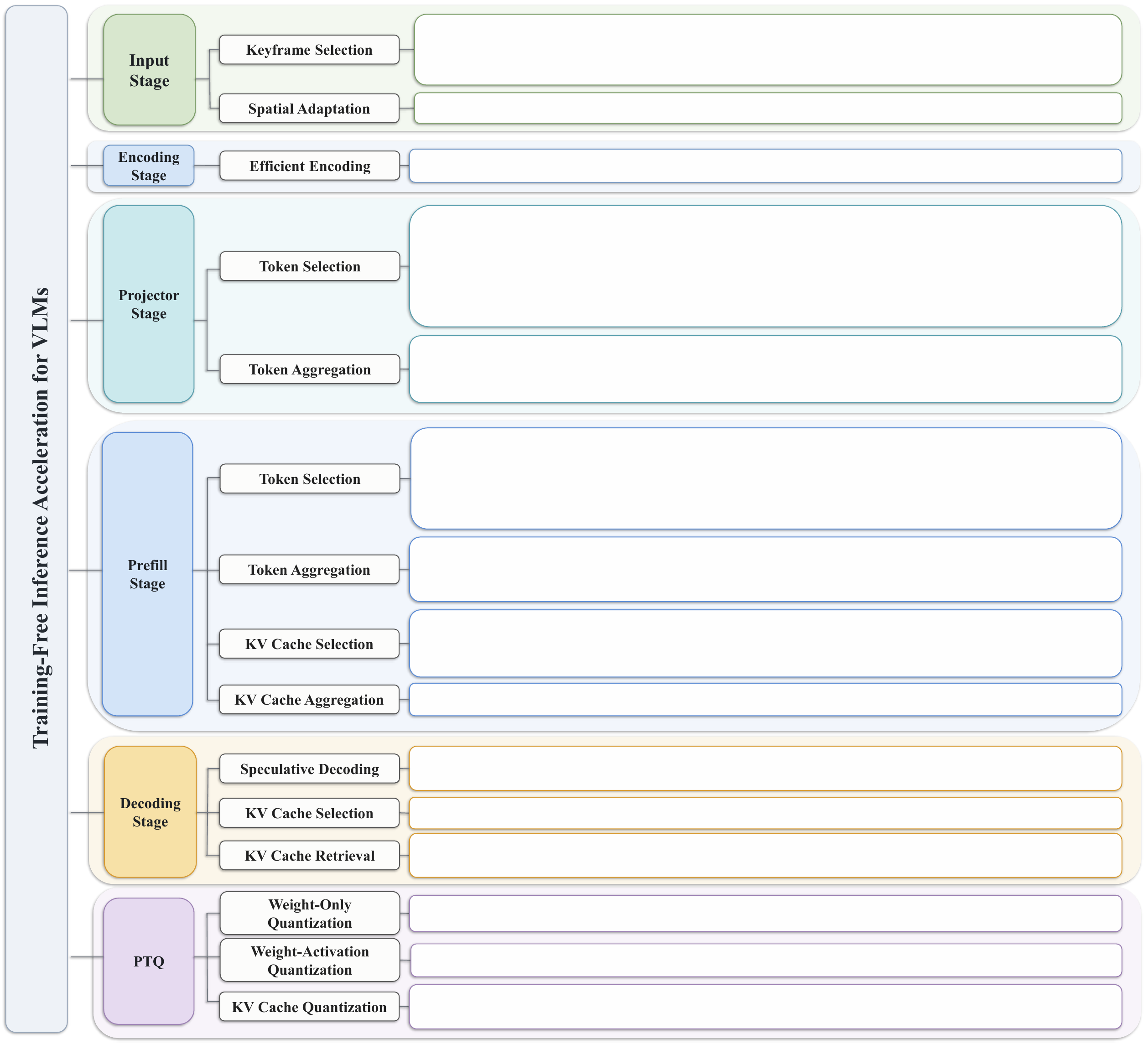}};
\begin{scope}[x={(image.south east)}, y={(image.north west)}]

\node[anchor=west, text width=0.61\textwidth, align=left,
      inner sep=0pt, font=\taxfont] at (0.365,0.9506){%
\textcolor{TaxGreen}{\textbf{Query-aware:}}~\paper{AKS}{tang2025adaptive}; \paper{BOLT}{liu2025bolt}; \paper{MDP\taxsup{3}}{sun2025mdp}; \paper{VideoTree}{wang2025videotree}; \paper{Q-Frame-QFS}{zhang2025q};  \paper{F2C}{sun2025frames}; \paper{FOCUS}{zhu2026focus}\\[0.8pt]
\textcolor{TaxGreen}{\textbf{Query-agnostic:}}~\paper{KTV-I}{song2026ktv}; \paper{MaxInfo}{li2026maxinfo};\paper{MEDR}{pu2026medr};
};

\node[anchor=west, text width=0.61\textwidth, align=left,
      inner sep=0pt, font=\taxfont] at (0.365,0.8961){%
\paper{Q-Frame-MRA}{zhang2025q}; \paper{F2C}{sun2025frames}
};

\node[anchor=west, text width=0.61\textwidth, align=left,
      inner sep=0pt, font=\taxfont] at (0.365,0.8400){%
\paper{EarlyTom-I}{wang2026earlytom}; \paper{ST-Merge}{chen2026fast}; \paper{STC-Cacher}{wang2026accelerating}
};

\node[anchor=west, text width=0.61\textwidth, align=left,
      inner sep=0pt, font=\taxfont] at (0.365,0.7440){%
\textcolor{TaxTeal}{\textbf{Attention-aware:}}~\paper{VisionZip}{yang2025visionzip}; \paper{LLaVA-PruMerge}{shang2025llava}; \paper{EarlyTom-II}{wang2026earlytom}; \paper{KTV-II}{song2026ktv}; \paper{StreamingTOM-CTR}{chen2026streamingtom}; \paper{Unified-Outer}{du2026unified}; \paper{FlashVID-Outer}{fan2026flashvid}\\[0.8pt]
\textcolor{TaxTeal}{\textbf{Attention-agnostic:}}~\paper{ReTaKe-DPSelect}{wang2024retake}; \paper{VidCom\taxsup{2}}{liu2025video}; \paper{DivPrune}{alvar2025divprune}; \paper{V-CAST}{lin2026v}; \paper{MMG-Vid}{ma2026mmg}; \paper{G-Prune}{jiang2025kind}; \paper{STC-Pruner}{wang2026accelerating}; \paper{FastVID}{shen2026fastvid}
};

\node[anchor=west, text width=0.61\textwidth, align=left,
      inner sep=0pt, font=\taxfont] at (0.365,0.6470){%
\paper{AIM-I}{zhong2025aim}; \paper{DyCoke-TTM}{tao2025dycoke}; \paper{PruneVid-I}{huang2025prunevid}; \paper{VisionZip}{yang2025visionzip}; \paper{LLaVA-PruMerge}{shang2025llava}; \paper{HoliTom-Outer}{shao2026holitom}; \paper{Unified-Outer}{du2026unified}; \paper{FlashVID-Outer}{fan2026flashvid}; \paper{SlowFast-LLaVA}{xu2024slowfast}
};

\node[anchor=west, text width=0.61\textwidth, align=left,
      inner sep=0pt, font=\taxfont] at (0.365,0.5400){%
\textcolor{TaxBlue}{\textbf{Attention-Based:}}~\paper{FastV}{chen2024image}; \paper{PyramidDrop}{xing2024pyramiddrop}; \paper{G-Search}{zhao2025accelerating}; \paper{AIM-II}{zhong2025aim}; \paper{AdaTP}{sun2025adatp}; \paper{PruneVid-II}{huang2025prunevid}; \paper{PriorTR}{chen2026accelerating}; \paper{FrameFusion}{fu2025framefusion}; \paper{SparseVLM}{zhang2024sparsevlm}; \paper{ZipVL-Attention}{he2024zipvl}\\[0.8pt]
\textcolor{TaxBlue}{\textbf{Similarity-Based:}}~\paper{DART}{wen2025stop}; \paper{SAINT}{jeddi2025similarity}
};

\node[anchor=west, text width=0.61\textwidth, align=left,
      inner sep=0pt, font=\taxfont] at (0.365,0.4525){%
\paper{SparseVLM}{zhang2024sparsevlm};
\paper{STTM}{hyun2025multi};
\paper{FrameFusion}{fu2025framefusion};
\paper{PACT}{dhouib2025pact};
\paper{Unified-Inner}{du2026unified};
\paper{HoliTom-Inner}{shao2026holitom}
};

\node[anchor=west, text width=0.61\textwidth, align=left,
      inner sep=0pt, font=\taxfont] at (0.365,0.3831){%
\paper{ReTaKe-PivotKV}{wang2024retake}; \paper{AirCache}{huang2025aircache}; \paper{ZipVL-Cache Construction}{he2024zipvl}; \paper{InfiniPot-V}{kim2026infinipot}; \paper{StreamMem}{yang2025streammem}; \paper{LOOK-M}{wan2024look}; \paper{MEDA}{wan2025meda}; \paper{LiveVLM-VSB}{ning2025livevlm}
};

\node[anchor=west, text width=0.61\textwidth, align=left,
      inner sep=0pt, font=\taxfont] at (0.365,0.3270){%
\paper{LOOK-M}{wan2024look}; \paper{MEDA}{wan2025meda}; \paper{StreamMem}{yang2025streammem}
};

\node[anchor=west, text width=0.61\textwidth, align=left,
      inner sep=0pt, font=\taxfont] at (0.365,0.2635){%
\paper{SpecVLM}{ji2025specvlm}; \paper{Sparse-to-Dense}{zhang2025sparse}; \paper{ParallelVLM}{kong2026parallelvlm}; \paper{LVSpec}{ji2026see}
};

\node[anchor=west, text width=0.61\textwidth, align=left,
      inner sep=0pt, font=\taxfont] at (0.365,0.2205){%
\paper{METok}{wang2025metok}; \paper{DyCoke-II}{tao2025dycoke}; \paper{VisCache}{wang2026viscache}
};

\node[anchor=west, text width=0.61\textwidth, align=left,
      inner sep=0pt, font=\taxfont] at (0.365,0.1785){%
\paper{SparseVILA}{khaki2025sparsevila}; \paper{ReKV}{di2025streaming}; \paper{LiveVLM-PaR}{ning2025livevlm}; \paper{StreamingTOM-OQM}{chen2026streamingtom}
};

\node[anchor=west, text width=0.61\textwidth, align=left,
      inner sep=0pt, font=\taxfont] at (0.365,0.1250){%
\paper{VLMQ}{xue2025vlmq}; \paper{MBQ}{li2025mbq}; \paper{VLM-PTQ}{deng2026vlm}; \paper{QIG}{xiang2026fine}
};

\node[anchor=west, text width=0.61\textwidth, align=left,
      inner sep=0pt, font=\taxfont] at (0.365,0.0783){%
\paper{Q-VLM}{wang2024q}; \paper{MBQ}{li2025mbq}; \paper{QIG}{xiang2026fine}; \paper{MASQuant}{hu2026masquant}
};

\node[anchor=west, text width=0.61\textwidth, align=left,
      inner sep=0pt, font=\taxfont] at (0.365,0.0326){%
\paper{CalibQuant}{han2025calibquant}; \paper{VidKV}{tao2025plug}; \paper{Dynamic KV Cache Quantization}{fan2026efficient}
};

\end{scope}
\end{tikzpicture}
\caption{Stage-wise taxonomy of training-free VLM
inference acceleration with representative methods. PTQ is included as a related cross-stage extension.}
\label{fig:taxonomy}
\end{figure*}

\section{Taxonomy}
\label{sec:taxonomy}

As shown in Fig.~\ref{fig:taxonomy}, we organize
training-free VLM acceleration methods by their
intervention stages in the inference pipeline:
\textit{Input} (Sec.~\ref{sec:input}),
\textit{Encoding} (Sec.~\ref{sec:encoding}),
\textit{Projector} (Sec.~\ref{sec:projector}),
\textit{Prefill} (Sec.~\ref{sec:prefill}), and
\textit{Decoding} (Sec.~\ref{sec:decoding}).
This stage-wise organization highlights differences
in computational savings and information availability.
We additionally review PTQ (Sec.~\ref{sec:ptq})
as a cross-stage extension, distinguishing its
calibration-dependent techniques from the core
training-free methods.

\subsection{Input Stage}
\label{sec:input}

Input-stage methods reduce or adapt visual content before
encoding, potentially saving both visual encoding and
downstream computation. They mainly operate through
\textit{keyframe selection}
(Sec.~\ref{sec:keyframe_selection}) and
\textit{spatial adaptation}
(Sec.~\ref{sec:spatial_adaptation}).

\subsubsection{Keyframe Selection}
\label{sec:keyframe_selection}

Reducing the frame budget lowers visual processing
costs but also limits downstream reasoning to the
retained temporal evidence.
Existing methods differ in whether selection is guided
by the query.

\noindent\textbf{Query-Agnostic.}
Uniform sampling provides a fixed-frame baseline
\cite{lin2024video,li2024mvbench,li2024llama}.
The first stage of KTV uses feature clustering
\cite{song2026ktv}, MaxInfo employs maximum-volume
selection \cite{li2026maxinfo}, and MEDR integrates
visual, motion, and text-derived event cues
\cite{pu2026medr}.
These methods support cross-query reuse but may
overlook question-specific evidence.

\noindent\textbf{Query-Aware.}
To better capture task-relevant evidence, BOLT exploits
query similarity \cite{liu2025bolt}, AKS balances
relevance and temporal coverage \cite{tang2025adaptive},
and MDP$^3$ considers diversity and temporal order
\cite{sun2025mdp}. The QFS component of Q-Frame employs query-guided
sampling \cite{zhang2025q}, while QCA combines relevance
with content diversity \cite{peng2026qca}.
VideoTree uses hierarchical retrieval
\cite{wang2025videotree}, whereas FOCUS adaptively
explores promising temporal regions \cite{zhu2026focus}. Greater query relevance may therefore come with
additional scoring or search costs, potentially
offsetting the savings from processing fewer frames.

\subsubsection{Spatial Adaptation}
\label{sec:spatial_adaptation}

Beyond temporal selection, resolution adjustment and
region-focused processing offer another way to reduce
visual computation. Free VLM employs prompt-guided cropping
\cite{han2024free}, Q-Frame's MRA component allocates
resolution by query relevance \cite{zhang2025q},
and F2C balances resolution and clip length
\cite{sun2025frames}. LENS further combines spatial
zoom-in with temporal zoom-out \cite{zhang2026lens}.
Reducing spatial computation, however, may obscure
fine-grained evidence or remove context needed
for accurate reasoning.

\subsection{Encoding Stage}
\label{sec:encoding}

Encoding-stage methods reduce visual computation by
merging or reusing intermediate features.
The first stage of EarlyTom merges temporally
redundant features in shallow encoder layers
\cite{wang2026earlytom}, ST-Merge performs
spatiotemporal token merging \cite{chen2026fast},
and STC-Cacher reuses stable features while
selectively updating dynamic content
\cite{wang2026accelerating}.
Early merging reduces subsequent encoding costs
but may obscure subtle motion, whereas caching
avoids repeated computation at the risk of
stale features and additional matching overhead.

\subsection{Projector Stage}
\label{sec:projector}

Projector-stage methods compress visual tokens after
encoding but before LLM processing, reducing prefill
costs without saving the encoding computation already
incurred. They mainly employ \textit{token selection}
(Sec.~\ref{sec:projector_selection}) and
\textit{token aggregation}
(Sec.~\ref{sec:projector_aggregation}).

\subsubsection{Token Selection}
\label{sec:projector_selection}

Selecting fewer visual tokens reduces LLM context length
but limits the evidence available for reasoning.
Existing approaches differ in whether they use attention
or feature-derived signals to estimate token importance.

\noindent\textbf{Attention-Aware.}
The second stage of EarlyTom uses encoder attention
for spatial selection \cite{wang2026earlytom},
while KTV's second stage combines [CLS] attention
with intra-frame similarity \cite{song2026ktv}. VisionZip retains dominant and
contextual tokens \cite{yang2025visionzip}, whereas
LLaVA-PruMerge combines attention-based selection
with feature merging \cite{shang2025llava}.
The CTR component of StreamingTOM selects tokens
using temporal changes and visual saliency
\cite{chen2026streamingtom}.
Such attention cues, however, may not align with
query-specific evidence.

\noindent\textbf{Attention-Agnostic.}
ReTaKe-DPSelect exploits temporal differences
\cite{wang2024retake}, while STC-Pruner evaluates
spatiotemporal token novelty
\cite{wang2026accelerating}. VidCom$^2$ uses frame
and token-level uniqueness \cite{liu2025video},
whereas V-CAST allocates temporal budgets based on
curvature and preserves representative spatial tokens
\cite{lin2026v}.
These feature-based criteria can improve coverage
but may retain content unrelated to the query.

\subsubsection{Token Aggregation}
\label{sec:projector_aggregation}

To retain more information from redundant tokens,
aggregation merges similar representations rather than
discarding them. AIM performs similarity-based merging
\cite{zhong2025aim}, PruneVid combines temporally
static and spatially similar tokens
\cite{huang2025prunevid}, and HoliTom employs global
temporal segmentation followed by spatiotemporal merging
\cite{shao2026holitom}.
While merging can preserve information lost by hard
selection, it may blur fine-grained distinctions and
introduce additional processing overhead.

\subsection{Prefill Stage}
\label{sec:prefill}

Prefill-stage methods exploit intermediate LLM
representations to reduce subsequent computation or
KV cache storage. We categorize them into
\textit{token selection} (Sec.~\ref{sec:prefill_selection}),
\textit{token aggregation} (Sec.~\ref{sec:prefill_aggregation}),
\textit{KV cache selection} (Sec.~\ref{sec:prefill_kv_selection}),
and \textit{KV cache aggregation}
(Sec.~\ref{sec:prefill_kv_aggregation}).

\subsubsection{Token Selection}
\label{sec:prefill_selection}

Pruning visual tokens within the LLM reduces computation
in subsequent layers, although earlier layers must still
process the original sequence. Methods mainly use
\textit{attention-based} or \textit{similarity-based} signals.

\noindent\textbf{Attention-Based.}
FastV and the second stage of PruneVid remove low-attention
tokens \cite{chen2024image,huang2025prunevid}, while
AIM's second stage propagates token importance through
attention \cite{zhong2025aim}. AdaTP corrects global
and local attention biases \cite{sun2025adatp}, whereas
PriorTR separates instruction relevance from model-induced
attention priors \cite{chen2026accelerating}.
Attention offers task-sensitive guidance, but biased
scores may discard critical visual evidence.

\noindent\textbf{Similarity-Based.}
DART removes tokens redundant with selected pivots
\cite{wen2025stop}, while SAINT applies similarity-aware
pruning \cite{jeddi2025similarity}.
Such criteria reduce reliance on attention scores,
although similar features may encode distinct events.

\subsubsection{Token Aggregation}
\label{sec:prefill_aggregation}

Rather than discarding visual tokens, aggregation merges
redundant representations to retain part of their
information. STTM performs multi-granular spatiotemporal
merging \cite{hyun2025multi}, while HoliTom's inner
stage combines attention and similarity
\cite{shao2026holitom}. Unified introduces text-aware
merging within the LLM \cite{du2026unified}, whereas
FrameFusion and PACT combine merging with pruning
\cite{fu2025framefusion,dhouib2025pact}.
Merging can preserve contextual information, but
may blur subtle event distinctions and incur
additional processing overhead.

\subsubsection{KV Cache Selection}
\label{sec:prefill_kv_selection}

Beyond token reduction, KV cache selection limits
the states retained for subsequent attention.
ReTaKe's PivotKV retains representative frame states
during chunked prefill \cite{wang2024retake}, while
InfiniPot-V and StreamMem maintain bounded historical
memories through selective retention
\cite{kim2026infinipot,yang2025streammem}.
This reduces cache pressure but risks losing
historical evidence needed by later queries.

\subsubsection{KV Cache Aggregation}
\label{sec:prefill_kv_aggregation}

To mitigate information loss from KV eviction,
LOOK-M merges selected cache states
\cite{wan2024look}, while MEDA combines KV merging
with layer-adaptive cache allocation
\cite{wan2025meda}. StreamMem further maintains
frame-level summaries alongside salient states
\cite{yang2025streammem}.
These compact representations reduce persistent
memory costs, but cannot recover the prefill
computation already incurred and may introduce
additional merging overhead.

\subsection{Decoding Stage}
\label{sec:decoding}

Autoregressive decoding repeatedly accesses accumulated
KV states, making time per output token (TPOT) sensitive
to long visual contexts. Training-free methods mainly
employ \textit{speculative decoding}
(Sec.~\ref{sec:speculative_decoding}),
\textit{KV cache selection}
(Sec.~\ref{sec:decoding_kv_selection}), and
\textit{KV cache retrieval}
(Sec.~\ref{sec:decoding_kv_retrieval}).

\subsubsection{Speculative Decoding}
\label{sec:speculative_decoding}

Speculative decoding (SD)\footnote{
Appendix~\ref{app:speculative} provides a detailed
workload-aware analysis of speculative decoding.}
accelerates generation through draft-and-verify execution
\cite{leviathan2023fast,chen2023accelerating}.
SpecVLM reduces drafting costs through verifier-guided
visual pruning \cite{ji2025specvlm}, while
Sparse-to-Dense (StD) combines sparse drafting with
dense verification \cite{zhang2025sparse}.
ParallelVLM overlaps drafting and verification
\cite{kong2026parallelvlm}, whereas LVSpec relaxes
verification to increase token acceptance
\cite{ji2026see}.
Speedup depends on drafting cost, acceptance, and
verification overhead, while relaxed verification
may sacrifice exact output fidelity.

\subsubsection{KV Cache Selection}
\label{sec:decoding_kv_selection}

Reducing the visual KV states accessed during generation
can lower repeated attention costs
\cite{tu2025vl}.
METok removes redundant visual KV states during
decoding \cite{wang2025metok}, while the second
stage of DyCoke dynamically updates the active cache
according to generation-time attention
\cite{tao2025dycoke}.
Such selection reduces cache access but may discard
evidence needed by subsequent tokens.

\subsubsection{KV Cache Retrieval}
\label{sec:decoding_kv_retrieval}

Rather than permanently discarding historical states,
retrieval preserves broader video context for selective
access. SparseVILA uses query-aware visual KV access
\cite{khaki2025sparsevila}, while ReKV retrieves
relevant states from stored video history
\cite{di2025streaming}.
LiveVLM's PaR enables page-level retrieval
\cite{ning2025livevlm}, whereas StreamingTOM's OQM
retrieves and dequantizes selected KV groups
\cite{chen2026streamingtom}.
Such access preserves long-range evidence but adds
retrieval overhead, with latency benefits depending
on when states are loaded.

\begin{table*}[t]
\centering
\scriptsize
\setlength{\tabcolsep}{3pt}
\renewcommand{\arraystretch}{1.05}

\resizebox{\textwidth}{!}{%
\begin{tabular}{@{}ll*{4}{cc}c@{}}
\toprule

\multirow{2}{*}{\textbf{Method}} &
\multirow{2}{*}{\textbf{Stage}} &
\multicolumn{2}{c}{\textbf{Video-MME}} &
\multicolumn{2}{c}{\textbf{EgoSchema}} &
\multicolumn{2}{c}{\textbf{LongVideoBench}} &
\multicolumn{2}{c}{\textbf{MVBench}} &
\multirow{2}{*}{\textbf{QLB$_{\mathrm{TTFT}}\uparrow$}} \\

\cmidrule(lr){3-4}
\cmidrule(lr){5-6}
\cmidrule(lr){7-8}
\cmidrule(lr){9-10}

& &
Acc.$\uparrow$ & TTFT$\downarrow$ &
Acc.$\uparrow$ & TTFT$\downarrow$ &
Acc.$\uparrow$ & TTFT$\downarrow$ &
Acc.$\uparrow$ & TTFT$\downarrow$ &
\\

\midrule

Full input & --
& 58.48 & 294
& 64.20 & 298
& 55.20 & 301
& 58.39 & 294
& 100.0 \\

Curv. merge & Proj.
& 57.89 & 236
& 64.60 & 239
& 54.76 & 240
& 58.03 & 236
& 110.6 \\

PriorTR & Pref.
& 57.70 & 294
& 63.00 & 296
& 54.27 & 291
& 57.89 & 255
& 100.7 \\

EarlyTom & Enc.+Proj.
& 57.85 & 160
& 62.80 & 163
& 54.00 & 160
& 56.08 & 161
& 126.6 \\

FrameFusion & Pref.
& 57.85 & 342
& 63.40 & 386
& 54.93 & 337
& 58.39 & 275
& 93.3 \\

VidCom$^2$ & Proj.
& 57.48 & 169
& 64.20 & 173
& 53.84 & 169
& 56.29 & 169
& 124.8 \\

FastVID & Proj.
& 56.78 & 187
& 64.60 & 183
& 53.20 & 189
& 56.50 & 169
& 121.2 \\

HoliTom & Proj.+Pref.
& 56.70 & 214
& 64.00 & 219
& 52.68 & 220
& 57.47 & 234
& 111.8 \\

AIM & Proj.+Pref.
& 56.78 & 258
& 63.20 & 272
& 52.32 & 271
& 56.24 & 255
& 102.4 \\

\bottomrule
\end{tabular}%
}

\caption{
Quality--latency comparison. Stage denotes the compression location:
Enc. (Encoding), Proj. (Projector), and Pref. (Prefill).
Accuracy is in \% and TTFT in ms.
Higher accuracy and QLB are better; lower TTFT is better.
}
\label{tab:quality_ttft}
\end{table*}

\subsection{Post-Training Quantization}
\label{sec:ptq}

Post-training quantization (PTQ) reduces the precision
of weights, activations, or KV states to lower memory
demands and potentially accelerate inference.
As a cross-stage extension, PTQ may involve calibration
or optimization beyond the core training-free scope.
We categorize methods into \textit{weight-only quantization}
(Sec.~\ref{sec:ptq_weight}),
\textit{weight-activation quantization}
(Sec.~\ref{sec:ptq_wa}), and
\textit{KV cache quantization}
(Sec.~\ref{sec:ptq_kv}).

\subsubsection{Weight-Only Quantization}
\label{sec:ptq_weight}

Quantizing weights reduces model storage and memory
traffic while retaining higher-precision activations.
VLMQ incorporates token importance into Hessian-based
quantization \cite{xue2025vlmq}, while VLM-PTQ
introduces modality-aware channel importance into
weight compensation \cite{deng2026vlm}.
MBQ balances vision and language sensitivities during
calibration \cite{li2025mbq}, whereas QIG estimates
fine-grained token sensitivity using integrated
gradients \cite{xiang2026fine}.
Lower-bit weights reduce memory pressure, but realized
speedups depend on kernel support and dequantization costs.

\subsubsection{Weight-Activation Quantization}
\label{sec:ptq_wa}

Further quantizing activations enables low-precision
computation but increases sensitivity to quantization
errors. Q-VLM considers cross-layer dependencies
during quantization search \cite{wang2024q},
while MBQ balances modality-specific sensitivities
\cite{li2025mbq} and QIG models token-level
interactions \cite{xiang2026fine}.
MASQuant introduces modality-aware smoothing and
cross-modal compensation \cite{hu2026masquant}.
The computational benefits depend on efficient
low-precision execution and sufficient accuracy retention.

\subsubsection{KV Cache Quantization}
\label{sec:ptq_kv}

Reducing KV precision lowers cache memory and bandwidth
demands, particularly for long visual contexts.
CalibQuant uses calibration and post-scaling for
1-bit visual KV quantization \cite{han2025calibquant},
while VidKV applies distinct low-bit strategies to
key and value caches in VLMs \cite{tao2025plug}.
Dynamic KV Cache Quantization adapts quantization
parameters to runtime KV distributions
\cite{fan2026efficient}.
Smaller caches, however, do not guarantee lower
latency when quantization and dequantization
overhead is considered.


\section{Evaluation}
\label{sec:evaluation}

Differences in model backbones, input budgets,
hardware, and measurement protocols limit direct
cross-study comparisons. We therefore combine our
controlled experiments with results reported in
prior studies, prioritizing comparisons under
matched settings.\footnote{
Appendix~\ref{app:evaluation} details the evaluation
metrics, benchmarks, and comparison protocols.}

Our controlled experiments use LLaVA-OneVision-7B
\cite{li2024llavaonevision} with 32 input frames
on an NVIDIA H100 (80 GB) GPU. We evaluate accuracy
and TTFT on Video-MME \cite{fu2025video},
EgoSchema \cite{mangalam2023egoschema},
LongVideoBench \cite{wu2024longvideobench},
and MVBench \cite{li2024mvbench}.
For literature-based comparisons, we examine
decoding on Qwen2.5-VL-3B \cite{bai2025qwen25vl}
and LLaVA-OneVision-7B using EgoSchema and
VideoDC \cite{chai2025auroracap};
PTQ on LLaVA-OneVision-7B and Qwen2.5-VL-7B;
and end-to-end efficiency on Qwen3-VL-8B
\cite{bai2025qwen3vl}.

To summarize quality--latency trade-offs, we
introduce \textbf{Quality--Latency Balance (QLB)}:
\begin{equation}
\mathrm{QLB}_{x}
=
100
\times
\frac{A}{A_0}
\times
\frac{2L_{0,x}}{L_{0,x}+L_x}.
\label{eq:qlb}
\end{equation}
Here, $A$ and $A_0$ denote the task quality of
an accelerated method and its matched baseline,
while $L_x$ and $L_{0,x}$ denote their measured
latency, with $x\in\{\mathrm{TTFT},\mathrm{TPOT}\}$.
For our controlled experiments, accuracy and TTFT
are averaged equally across the four benchmarks.
Each matched baseline scores 100, with higher QLB
indicating a more favorable relative quality--latency
balance.

\noindent\textbf{Research Questions.}
We investigate quality--latency trade-offs across
two inference phases: how intervention position and
compression strategy affect accuracy and pre-generation
latency (RQ1), and how decoding acceleration balances
generation speed and output quality (RQ2).
We further examine how well PTQ preserves task quality
under reduced numerical precision (RQ3), and whether
local efficiency gains translate into end-to-end
acceleration (RQ4).

\begin{table}[t]
\centering
\footnotesize
\setlength{\tabcolsep}{3.0pt}
\renewcommand{\arraystretch}{1.05}

\begin{tabular*}{\columnwidth}
{@{\extracolsep{\fill}}lcccc@{}}
\toprule
\textbf{Method} &
\textbf{Ret.} &
\textbf{Quality$\uparrow$} &
\textbf{TPOT$\downarrow$} &
\textbf{QLB$_{\mathrm{TPOT}}\uparrow$} \\
\midrule

\multicolumn{5}{@{}l}{
\textit{EgoSchema: Qwen2.5-VL-3B, Acc. (\%)}} \\

Full KV     & 100\% & 57.20 & 59.37 & 100.0 \\
VisCache    & 40\%  & 52.00 & 56.82 & 92.9  \\
VisCache    & 28\%  & 55.00 & 49.50 & 104.9 \\
VisCache    & 19\%  & 54.60 & 47.70 & 105.9 \\

\midrule

\multicolumn{5}{@{}l}{
\textit{VideoDC: LLaVA-OneVision-7B, VDCscore}} \\

Full tokens & Full  & 3.30 & 42 & 100.0 \\
DyCoke & 18.75\% & 3.29 & 35 & 108.8 \\
DyCoke & 14.25\% & 3.20 & 31 & 111.6 \\
\bottomrule
\end{tabular*}

\caption{
Quality--latency trade-offs during decoding.
Reported hardware includes 4$\times$A100 (80 GB)
for VisCache and A6000 (48 GB) for DyCoke.
Ret. denotes KV-cache or visual-token retention (\%);
Quality reports EgoSchema accuracy (\%) or VideoDC
captioning score, and TPOT is in ms/token.
}
\label{tab:decoding}
\end{table}

\noindent\textbf{RQ1.}
Table~\ref{tab:quality_ttft} highlights how intervention
position shapes the balance between computational savings
and evidence preservation. EarlyTom \cite{wang2026earlytom}
reduces mean TTFT by 45.7\% while retaining 97.7\% of
baseline accuracy (QLB 126.6). In comparison,
Curv. merge\footnote{
A curvature-guided token-merging baseline applied
after visual encoding and before LLM prefill.}
preserves more accuracy (99.6\%) but achieves a smaller
TTFT reduction (19.9\%). Yet post-encoder compression
can remain competitive, as VidCom$^2$ \cite{liu2025video}
reaches a QLB of 124.8. Conversely, FrameFusion
\cite{fu2025framefusion} increases TTFT by 12.9\%
despite token reduction. These results suggest that
earlier compression offers greater opportunities to
avoid downstream computation, while evidence preservation
and processing overhead determine the realized benefit. However, differences in retention budgets and
implementations prevent isolating intervention
position as the sole cause.

\noindent\textbf{RQ2.}
Table~\ref{tab:decoding} examines how reducing retained
visual representations affects decoding efficiency and
task quality. For VisCache \cite{wang2026viscache},
reducing retention from 40\% to 28\% lowers TPOT
from 56.82 to 49.50\,ms/token while improving
EgoSchema accuracy from 52.0\% to 55.0\%, raising
QLB from 92.9 to 104.9. This non-monotonic trend
suggests that retention ratio alone does not determine
the utility of preserved evidence. DyCoke
\cite{tao2025dycoke} further illustrates the trade-off:
TPOT decreases from 42 to 31\,ms/token, while the
VideoDC \cite{chai2025auroracap} score drops from
3.30 to 3.20 (QLB 111.6). Overall, decoding
acceleration depends not only on reducing token
processing but also on preserving information needed
for subsequent generation.

\noindent\textbf{RQ3.}
Table~\ref{tab:ptq} examines quality retention under
reduced numerical precision. On LLaVA-OneVision-7B
at W3A16, VLM-PTQ \cite{deng2026vlm} retains
97.3\% of FP16 performance, compared with
92.0\% for VLMQ \cite{xue2025vlmq} and
90.2\% for MBQ \cite{li2025mbq}.
In separate five-benchmark comparisons, QIG
\cite{xiang2026fine} improves retention from
96.0\% to 98.2\%, while MASQuant
\cite{hu2026masquant} increases it from
91.9\% to 92.6\% under W4A8.
These results indicate that bit-width alone
does not determine quality preservation.
However, these image-oriented evaluations provide
no matched latency or memory measurements,
leaving the realized acceleration benefits
for video inference unverified.

\begin{table}[t]
\centering
\footnotesize
\setlength{\tabcolsep}{3pt}
\renewcommand{\arraystretch}{1.05}

\begin{tabular*}{\columnwidth}
{@{\extracolsep{\fill}}lccc@{}}
\toprule
\textbf{Method} &
\textbf{Bits} &
\textbf{Avg. (Q/FP16)} &
\textbf{Q-ret.$\uparrow$} \\
\midrule

\multicolumn{4}{@{}l}{
\textit{LLaVA-OneVision-7B, 8 benchmarks}} \\
MBQ       & W3A16 & 64.4/71.4   & 90.2\% \\
VLMQ      & W3A16 & 65.7/71.4   & 92.0\% \\
VLM-PTQ   & W3A16 & 69.5/71.4   & 97.3\% \\

\midrule
\multicolumn{4}{@{}l}{
\textit{LLaVA-OneVision-7B, 5 benchmarks}} \\
MBQ       & W3A16 & 70.44/73.37 & 96.0\% \\
QIG       & W3A16 & 72.04/73.37 & 98.2\% \\

\midrule
\multicolumn{4}{@{}l}{
\textit{Qwen2.5-VL-7B, 5 benchmarks}} \\
MBQ       & W4A8  & 68.5/74.5   & 91.9\% \\
MASQuant  & W4A8  & 69.0/74.5   & 92.6\% \\

\bottomrule
\end{tabular*}

\caption{
PTQ quality retention under the reported model,
bit-width, and benchmark settings.
Q/FP16 denotes the average scores of quantized
and FP16 models; Q-ret. denotes their ratio (\%).
}
\label{tab:ptq}
\end{table}

\noindent\textbf{RQ4.}
Table~\ref{tab:e2e} presents a matched system-level
comparison reported by V-CAST \cite{lin2026v},
covering representative compression methods
\cite{yang2025visionzip,liu2025video,
shao2026holitom,fan2026flashvid}.
Although V-CAST reduces prefill and generation
latency by 51.2\% and 46.6\%, respectively,
end-to-end latency decreases by only 13.6\%,
with a 1.0-point accuracy loss.
Stage-level rankings can also differ from
end-to-end outcomes: VidCom$^2$ is faster than
HoliTom in both reported stages but has slightly
higher E2E latency. Moreover, FlashVID reduces
both stage latencies while increasing peak memory
by 84.0\%. These results show that local
computational savings do not necessarily yield
proportional system-level benefits.

\begin{table}[t]
\centering
\scriptsize
\setlength{\tabcolsep}{1.8pt}
\renewcommand{\arraystretch}{1.05}

\begin{tabular*}{\columnwidth}
{@{\extracolsep{\fill}}lrrrrr@{}}
\toprule
\textbf{Method} &
\textbf{Acc.$\uparrow$} &
\textbf{Prefill$\downarrow$} &
\textbf{Gen.$\downarrow$} &
\textbf{E2E$\downarrow$} &
\textbf{Peak Mem.$\downarrow$} \\
\midrule

Qwen3-VL-8B
& 64.5 & 243.6 & 280.9 & 1440.4 & 22478.0 \\

VisionZip
& 60.1 & 155.6 & 190.5 & 1276.8 & 19974.7 \\

VidCom$^2$
& 62.4 & 119.0 & 150.2 & 1271.0 & 19547.5 \\

HoliTom
& 59.7 & 134.3 & 160.8 & 1261.8 & 19974.7 \\

FlashVID
& 62.3 & 146.6 & 175.0 & 1308.7 & 41363.6 \\

V-CAST
& 63.5 & 118.8 & \textbf{149.9}
& 1245.1 & 19547.5 \\

\bottomrule
\end{tabular*}

\caption{
System-level quality and efficiency on Video-MME
using Qwen3-VL-8B-Instruct with up to 32 frames
and 25\% token retention for compressed methods.
Experiments use NVIDIA H100 (80 GB) GPUs. Prefill and Gen. denote the reported pre-generation
and autoregressive generation times, respectively. 
Accuracy is in \%, aggregate latency in seconds,
and peak GPU memory in MB.
}
\label{tab:e2e}
\end{table}

\section{Challenges and Future Work}
\label{sec:future}

Our taxonomy and evaluation reveal that training-free
acceleration depends not only on reducing visual tokens,
but also on preserving useful evidence and translating
local savings into actual inference gains. We identify
four directions for future research.

\noindent\textbf{Evidence-Aware Compression.}
Video redundancy is unevenly distributed, and brief
events may be lost under aggressive compression.
DToMA and RLP explore evidence recovery and semantic
preservation \cite{yuan2025dtoma,ma2026rlp}, while
DyToK and GTR adapt token budgets to frame importance
and query relevance \cite{li2025dytok,zhou2026gtr}.
D-CoDe further combines compression with question
decomposition \cite{huang2025dcode}.
Future methods should coordinate temporal coverage,
task relevance, and selective evidence recovery
without introducing excessive selection overhead.

\noindent\textbf{Early Encoder Optimization.}
Compared with post-encoder compression, training-free
optimization within video encoders remains relatively
underexplored. EarlyTom and STC demonstrate the
potential of early merging and feature reuse
\cite{wang2026earlytom,wang2026accelerating},
while MeToM exploits video metadata to reduce
redundant encoding \cite{wu2026metom}.
Encoder-aware selection in broader VLMs provides
further insights \cite{wang2026posprune,
liu2026hiprune}.
Future work should explore motion-aware feature
reuse and coordinated compression across encoding
and prefill, without discarding evidence required
by downstream reasoning.

\noindent\textbf{Streaming Memory and Decoding.}
Streaming VLMs must preserve historical evidence
under bounded memory while processing frames causally.
StreamingTOM and MuKV explore compressed visual
memory \cite{chen2026streamingtom,xiao2026mukv},
whereas StreamMeCo and StreamKV investigate
persistent memory compression and selective retrieval
\cite{wang2026streammeco,chen2026streamkv}.
Future systems should jointly manage retention,
retrieval, and cache updates while maintaining
timely responses to incoming queries
\cite{vasu2026vsas}.

\noindent\textbf{End-to-End System Co-Design.}
As shown in Table~\ref{tab:e2e}, substantial
stage-level savings may yield limited end-to-end
improvements. Token selection, memory access, and
kernel inefficiency can offset computational gains
\cite{li2026vista,ma2026apet}.
Future research should coordinate token compression,
KV management, and low-precision execution,
optimizing measured latency and memory alongside
task quality rather than compression ratios alone.

Taken together, these directions call for a shift
from isolated compression strategies to coordinated,
evidence-aware inference. The ultimate goal is not
merely to reduce visual tokens, but to preserve
reliable video reasoning while delivering measurable
end-to-end efficiency gains.

\section{Conclusion}

This survey highlights a central insight: training-free
VLM acceleration is not merely about reducing visual
tokens, but about balancing computational savings
with the preservation of task-relevant spatiotemporal
evidence. Across the inference pipeline, intervention
position determines both which computations can be
avoided and what information remains available for
downstream reasoning. Earlier compression offers
broader computational savings but risks irreversible
evidence loss, whereas later intervention exploits
richer representations while inheriting upstream costs.
Our analysis further reveals that reduced token counts
and local latency improvements do not necessarily
translate into end-to-end efficiency. These findings
point toward a shift from isolated compression
strategies to evidence-aware, cross-stage optimization,
where computation and memory are allocated according
to task demands and actual system bottlenecks.
Ultimately, progress should be measured by reliable
video understanding and realized inference efficiency.

\section*{Limitations}

This survey focuses on publicly available
training-free VLM acceleration methods and may
overlook recent or proprietary advances.
Our controlled experiments use a single backbone,
a fixed frame budget, and one GPU platform,
limiting their generalizability.
Other comparisons rely on published results
with potentially different experimental protocols.
Moreover, PTQ evidence is largely image-based,
and available system measurements lack complete
latency breakdowns.
Therefore, our findings should be interpreted
within their respective evaluation settings
rather than as universal performance rankings.

\section*{Ethical Considerations}

This survey synthesizes existing literature and
involves no human participants. Training-free
acceleration may reduce computational costs and
improve accessibility to video-LLMs. However,
lossy compression may discard critical evidence
and increase unreliable predictions. We encourage
the community to adopt robust evaluation protocols
that assess these risks alongside accuracy and
latency. Our experiments use publicly available
models and benchmarks, subject to their respective
licenses and usage terms. We also document
evaluation settings and comparison protocols
to support reproducibility.


\bibliography{custom}

\appendix

\begin{table*}[t]
\centering
\footnotesize
\setlength{\tabcolsep}{3.5pt}
\renewcommand{\arraystretch}{1.10}

\begin{tabular}{@{}lp{4.3cm}lp{6.7cm}@{}}
\toprule
\textbf{Metric} &
\textbf{Formulation} &
\textbf{Scope} &
\textbf{Definition} \\
\midrule

\multicolumn{4}{@{}l}{
\textit{Task Quality and Compression}} \\

Accuracy
& $\displaystyle
\frac{N_{\mathrm{correct}}}
{N_{\mathrm{total}}}\times100\%$
& Quality
& Percentage of correctly answered questions
under the corresponding benchmark protocol. \\

VDCscore
& Benchmark-reported score
& Quality
& Video captioning quality measured using
the VideoDC evaluation protocol. \\

Ret.
& $\displaystyle
\frac{N_{\mathrm{retained}}}
{N_{\mathrm{original}}}\times100\%$
& Compression
& Percentage of retained visual tokens or
KV states relative to the source-defined budget. \\

Q-ret.
& $\displaystyle
\frac{Q_{\mathrm{quant}}}
{Q_{\mathrm{FP16}}}\times100\%$
& PTQ
& Quantized-model quality relative to the
matched FP16 baseline. \\

Bits
& $\mathrm{W}b_w\mathrm{A}b_a$
& PTQ
& Numerical precision of weights and
activations; KV-cache precision is specified
separately when applicable. \\

\midrule
\multicolumn{4}{@{}l}{
\textit{Inference Efficiency}} \\

TTFT
& $\displaystyle
t_{\mathrm{first}}-t_{\mathrm{arr}}$
& Pre-generation
& Elapsed time from request arrival to the
first output token under the stated timing boundary. \\

TPOT
& $\displaystyle
\frac{t_{\mathrm{end}}-t_{\mathrm{first}}}
{N_o-1}$
& Decoding
& Average generation latency per output token
after the first token. \\

E2E
& $\displaystyle
t_{\mathrm{end}}-t_{\mathrm{arr}}$
& Full Pipeline
& Total elapsed time from request arrival to
response completion. \\

Peak Mem.
& $\displaystyle
\max_t M_{\mathrm{GPU}}(t)$
& Full Pipeline
& Maximum GPU memory usage within the
specified measurement window. \\

\midrule
\multicolumn{4}{@{}l}{
\textit{Quality--Latency Evaluation}} \\

QLB$_x$
& $\displaystyle
100\times\frac{A}{A_0}\times
\frac{2L_{0,x}}{L_{0,x}+L_x}$
& Joint
& Baseline-normalized quality--latency score,
where $x\in\{\mathrm{TTFT},\mathrm{TPOT}\}$. \\

\bottomrule
\end{tabular}

\caption{
Core metrics for evaluating task quality,
compression, and realized inference efficiency.
QLB measures the relative quality--latency
balance under matched evaluation settings.
}
\label{tab:efficiency_metrics}
\end{table*}
\section{Evaluation Metrics and Evidence Protocol}
\label{app:evaluation}

\subsection{General Evaluation Metrics}
\label{app:metrics}

We evaluate training-free VLM acceleration along three
dimensions: \textit{task quality}, \textit{compression},
and \textit{realized inference cost}.
Task quality is assessed using benchmark accuracy or
task-specific scores, while compression is characterized
by visual-token or KV-cache retention and numerical
precision. Inference efficiency is measured using
Time to First Token (TTFT), Time per Output Token
(TPOT), End-to-End latency (E2E), and Peak Memory.
We further employ the baseline-normalized
Quality--Latency Balance (QLB) to assess the trade-off
between task quality and latency under matched settings.
Table~\ref{tab:efficiency_metrics} summarizes the
metrics and their definitions.

Here, $t_{\mathrm{arr}}$, $t_{\mathrm{first}}$, and
$t_{\mathrm{end}}$ denote request arrival, first-token
generation, and response completion times, respectively,
and $N_o$ is the number of generated tokens ($N_o>1$
for TPOT). For QLB, $A$ and $A_0$ denote the
task-specific quality scores of the accelerated method
and its matched baseline, while $L_x$ and $L_{0,x}$
denote their corresponding latencies.
The matched baseline has a QLB of 100.
Higher values indicate a more favorable relative
quality--latency balance, but do not imply
comparability across different evaluation settings.

\subsection{Video Understanding Benchmarks}
\label{app:benchmarks}

Table~\ref{tab:video_benchmarks} summarizes
representative benchmarks for video understanding,
covering temporal reasoning, egocentric understanding,
long-context comprehension, motion perception,
and detailed video description.

Our controlled evaluation uses Video-MME,
EgoSchema, LongVideoBench, and MVBench
\cite{fu2025video,mangalam2023egoschema,
wu2024longvideobench,li2024mvbench}.
The literature-based decoding comparison
additionally includes Video Detailed Captioning
(VideoDC/VDC), which evaluates generated
descriptions using VDCscore \cite{chai2025auroracap}.
Other benchmarks provide complementary coverage
of temporal perception, long-form reasoning,
and domain-specific understanding
\cite{yu2019activitynet,xiao2021next,
patraucean2023perception,liu2024tempcompass,
fang2024mmbench,song2024moviechat,
zhou2025mlvu,wang2025lvbench,
hong2025motionbench,hu2026video}.
These benchmarks differ in task format, video
duration, and evaluation protocol; results should
therefore be compared only under matched settings.

\begin{table*}[t]
\centering
\footnotesize
\setlength{\tabcolsep}{3.5pt}
\renewcommand{\arraystretch}{1.08}

\begin{tabular}{@{}lccp{2.1cm}p{3.0cm}p{5.0cm}@{}}
\toprule
\textbf{Benchmark} &
\textbf{Venue} &
\textbf{Year} &
\textbf{Video Length} &
\textbf{Scale} &
\textbf{Primary Focus} \\
\midrule

\multicolumn{6}{@{}l}{
\textit{Benchmarks used in quantitative comparisons}} \\

Video-MME
& CVPR & 2025
& 11 s--1 h
& 900 videos, 2.7K QA
& General video understanding across durations \\

EgoSchema
& NeurIPS & 2023
& Avg. 180 s
& 5.0K videos, 5.0K QA
& Long-horizon egocentric reasoning \\

LongVideoBench
& NeurIPS & 2024
& Up to 1 h
& 3.8K videos, 6.7K QA
& Long-context referring and temporal reasoning \\

MVBench
& CVPR & 2024
& Avg. 16 s
& 3.6K videos, 4.0K QA
& Multi-task temporal understanding \\

VideoDC (VDC)
& ICLR & 2025
& --
& $\sim$1.0K clips, captions
& Detailed video captioning and VDCscore \\

\midrule

\multicolumn{6}{@{}l}{
\textit{Additional representative video benchmarks}} \\

ActivityNet-QA
& AAAI & 2019
& Avg. 180 s
& 5.8K videos, 58K QA
& Long-form open-ended video QA \\

NExT-QA
& CVPR & 2021
& Avg. 44 s
& 5.4K videos, 52K QA
& Causal and temporal reasoning \\

Perception Test
& NeurIPS & 2023
& Avg. 23 s
& 11.6K videos, 44K QA
& Multimodal perception and reasoning \\

TempCompass
& ACL Findings & 2024
& --
& 410 videos, 7.54K tasks
& Fine-grained temporal perception \\

MMBench-Video
& NeurIPS & 2024
& 30 s--6 min
& 609 videos, 2.0K QA
& Holistic multi-shot video understanding \\

MovieChat-1K
& CVPR & 2024
& Avg. 9.4 min
& 1K videos, 13K QA
& Long-video dialogue and event understanding \\

MLVU
& CVPR & 2025
& Avg. 15.5 min
& 1.7K videos, 3.1K QA
& Multi-task long-video understanding \\

LVBench
& ICCV & 2025
& Avg. 4101 s
& 103 videos, 1.5K QA
& Extremely long-video understanding \\

MotionBench
& CVPR & 2025
& --
& 5.4K videos, 8.1K QA
& Fine-grained motion perception \\

Video-MMMU
& ACL & 2026
& Avg. 506 s
& 300 videos, 900 QA
& Knowledge acquisition from professional videos \\

\bottomrule
\end{tabular}

\caption{
Representative benchmarks for video understanding.
The first group includes datasets used in the
main-text quantitative comparisons; the second
provides broader task coverage.
Dataset sizes are approximate and may vary
across evaluation subsets.
}
\label{tab:video_benchmarks}
\end{table*}


\subsection{Host VLMs and Language Backbones}
\label{app:backbones}

The choice of host VLM plays an important role in
evaluating training-free inference acceleration.
Differences in visual tokenization, context capacity,
and model architecture affect both baseline performance
and computational cost. Consequently, similar retention
budgets may correspond to different inference workloads
across host models.

Existing studies employ diverse VLM backbones,
including Video-LLaVA, LLaVA-OneVision, and the
Qwen-VL family \cite{lin2024video,wang2024qwen2vl,
bai2025qwen25vl,bai2025qwen3vl}.
Table~\ref{tab:training_free_benchmark} summarizes
representative methods, their host models, compression
budgets, and reported performance across video
understanding benchmarks.
The table preserves a broad view of the literature,
including stage-specific components, multi-stage
methods, and architecture-dependent approaches
\cite{liu2025bolt,wang2026earlytom,
liu2025video,lin2026v,wang2026viscache}.
Since these results originate from different models,
budgets, and evaluation protocols, they serve as
reported evidence rather than a direct cross-method
ranking.

\begin{table*}[t]
\centering
\scriptsize
\setlength{\tabcolsep}{2.8pt}
\renewcommand{\arraystretch}{1.08}

\begin{tabular*}{\textwidth}{
@{\extracolsep{\fill}}
lllcccccc@{}
}
\toprule
\textbf{Method} &
\textbf{Host VLM} &
\textbf{Ret. / Budget} &
\textbf{NExT} &
\textbf{MVB} &
\textbf{V-MME} &
\textbf{ES} &
\textbf{LVB} &
\textbf{MLVU} \\
\midrule

\rowcolor{black!10}
\multicolumn{9}{c}{\textit{\textbf{Input Stage}}} \\
\midrule

AKS
& LLaVA-Video-7B
& 64f
& -- & -- & 65.3 & -- & 62.7 & -- \\

BOLT
& LLaVA-OneVision-7B
& 1fps $\rightarrow$ 8f
& 77.4 & -- & 56.1 & 59.23 & 54.5 & 63.6 \\

MDP$^{3}$
& LLaVA-OneVision-7B
& 128 $\rightarrow$ 8f
& -- & -- & 59.6 & -- & 59.0 & 69.8 \\

VideoTree
& GPT-4
& Adaptive
& 75.6 & -- & -- & 61.1 & -- & -- \\

Q-Frame
& Qwen2-VL-7B
& 4H+8M+32L
& -- & -- & 58.3 & -- & 58.4 & 65.4 \\

F2C
& Qwen2.5-VL-7B
& 8f-equiv.
& -- & -- & 61.4 & -- & 58.9 & 63.1 \\

FOCUS
& LLaVA-Video-7B
& 64f
& -- & -- & 65.4 & -- & 63.5 & -- \\

KTV$^{\dagger}$
& LLaVA-v1.6-7B
& 504 vis. tok.
& 64.5 & 46.2 & 43.6 & 49.6 & -- & 44.8 \\

MaxInfo
& LLaVA-Video-7B
& Adaptive ($\leq$64f)
& -- & -- & 64.2 & 63.7 & 61.5 & -- \\

\midrule

\rowcolor{black!10}
\multicolumn{9}{c}{\textit{\textbf{Encoding Stage}}} \\
\midrule

EarlyTom-I
& LLaVA-OV-7B
& 73.9\% avg. tok.
& -- & 57.9 & 57.0 & 60.3 & -- & -- \\

STC-Cacher
& StreamForest / Qwen2-7B
& 75\% cache reuse
& -- & -- & 60.7 & 59.7 & -- & 68.1 \\

\midrule

\rowcolor{black!10}
\multicolumn{9}{c}{\textit{\textbf{Projector Stage}}} \\
\midrule

EarlyTom-II$^{\dagger}$
& LLaVA-OV-7B
& 25\%
& -- & 57.4 & 58.5 & 60.5 & 56.3 & -- \\

KTV-II$^{\dagger}$
& LLaVA-v1.6-7B
& 936 tok.
& 65.1 & 46.4 & 43.7 & 50.4 & -- & 36.1 \\

StreamingTOM-CTR$^{\dagger}$
& LLaVA-OV-7B
& 50 tok./frame
& -- & -- & 59.9 & 63.7 & -- & 67.9 \\

Unified-Outer
& LLaVA-OV-7B
& 10\%
& -- & 57.4 & 56.6 & 60.5 & 57.4 & 44.7 \\

FlashVID-Outer
& LLaVA-OV-7B
& 10\%
& -- & 57.4 & 57.8 & 60.0 & 56.5 & -- \\

VidCom$^{2}$
& LLaVA-OV-7B
& 25\%
& -- & 57.2 & 58.6 & -- & 54.9 & 62.5 \\

DivPrune
& LLaVA-NeXT-Video
& --
& 24.48$^{*}$ & -- & -- & 39.76 & -- & -- \\

V-CAST
& LLaVA-OV-7B
& 25\%
& -- & 57.4 & 58.6 & -- & 56.4 & 62.9 \\

MMG-Vid
& LLaVA-OV-7B
& 25\%
& -- & 56.7 & 58.6 & -- & 56.6 & 62.9 \\

STC-Pruner
& ReKV / LLaVA-OV-7B
& 25\%
& -- & -- & 57.1 & 60.8 & -- & 67.6 \\

FastVID
& LLaVA-OV-7B
& 10\%
& -- & 55.9 & 57.3 & -- & 56.3 & 62.7 \\

VisionZip
& Qwen3-VL-8B
& 25\% tok.
& -- & -- & 60.1 & -- & -- & -- \\

AIM-I
& LLaVA-OV-7B
& 25\%
& -- & -- & 58.0 & -- & -- & -- \\

DyCoke-TTM
& LLaVA-OV-7B
& 14.25\%
& 77.2 & -- & 58.1 & -- & -- & -- \\

PruneVid-Stage I
& PLLaVA-7B
& 36.1\%
& -- & 47.2 & 44.7 & -- & -- & -- \\

HoliTom-Outer
& LLaVA-OV-7B
& 15\%
& -- & 58.1 & 58.1 & 61.0 & 57.0 & -- \\

SlowFast-LLaVA
& LLaVA-v1.6-Vicuna-7B
& 3680 tok.
& 64.2 & -- & -- & 47.2 & -- & -- \\

\midrule

\rowcolor{black!10}
\multicolumn{9}{c}{\textit{\textbf{Prefill Stage}}} \\
\midrule

AIM-II$^{\dagger}$
& LLaVA-OneVision-7B
& 25\% pre-LLM tok.
& -- & 57.1 & 58.2 & 59.6 & -- & 63.7 \\

AdaTP
& LLaVA-OneVision-7B
& 40$\rightarrow$12\% tok.
& -- & -- & 59.04 & -- & 56.02 & 63.47 \\

PruneVid-II$^{\dagger}$
& LLaVA-OneVision-7B
& 17.0\% tok.
& -- & 57.5 & 58.6 & 59.5 & -- & -- \\

FrameFusion
& LLaVA-Video-7B
& 30\% tok.
& 81.8 & 59.7 & 61.3 & 53.0 & -- & -- \\

STTM
& LLaVA-Video-7B
& 50\% tok.
& 82.5 & -- & 62.6 & 58.6 & 59.6 & 69.9 \\

PACT
& LLaVA-OneVision-7B
& $\approx$30.8\% tok.
& -- & -- & 57.7 & 60.0 & -- & 64.8 \\

Unified-Inner$^{\dagger}$
& LLaVA-OneVision-7B
& 10\%$\rightarrow$5\% tok.
& -- & 57.7 & 56.6 & 60.6 & 56.4 & 45.5 \\

HoliTom-Inner$^{\dagger}$
& LLaVA-OneVision-7B
& 10\%$\rightarrow$5\% tok.
& -- & 57.3 & 56.8 & 61.2 & 56.3 & -- \\

ReTaKe-PivotKV$^{\dagger}$
& Qwen2-VL-7B
& 2048f / $\leq$32K ctx.
& -- & -- & 63.9 & -- & 57.7 & 69.8 \\

InfiniPot-V
& Qwen2-VL-7B
& 768f / 6K KV
& -- & -- & 62.8 & 65.6 & 58.4 & 65.8 \\

StreamMem
& Qwen2-VL-7B
& 4.0/0.5 fps / 6K KV
& -- & -- & 62.1 & 67.2 & -- & 65.9 \\

LiveVLM-VSB$^{\dagger}$
& LLaVA-OneVision-7B
& 12K cache
& -- & -- & 59.6 & -- & 56.1 & 68.1 \\

\midrule

\rowcolor{black!10}
\multicolumn{9}{c}{\textit{\textbf{Decoding Stage}}} \\
\midrule

METok$^{\dagger}$
& LLaVA-OV-7B
& Multi-stage
& -- & 57.3 & 58.4 & 59.8 & -- & 64.7 \\

DyCoke-II$^{\dagger}$
& LLaVA-OV-7B
& 25\% before-LLM budget
& -- & 53.1 & 54.3 & 59.5 & 49.5 & -- \\

SparseVILA
& LongVILA-Qwen2-7B
& 50\% prefill / 25\% decode KV
& -- & -- & 59.8 & -- & -- & 72.9 \\

ReKV
& LLaVA-OV-7B
& 0.5 FPS $\rightarrow$ 64 retrieved frames
& -- & -- & -- & 60.7 & -- & 68.5 \\

LiveVLM-PaR
& LLaVA-OV-7B
& Streaming KV / retrieval
& -- & -- & 57.3 & 63.0 & -- & 66.3 \\

StreamingTOM-OQM$^{\dagger}$
& LLaVA-OV-7B
& 4-bit OQM, $\sim$12K retrieval
& -- & -- & 59.9 & 63.7 & -- & 67.9 \\

VisCache$^{\mathrm{C}}$
& Qwen2.5-VL-3B
& 28\% KV
& -- & -- & -- & 55.0 & -- & -- \\

\bottomrule
\end{tabular*}

\caption{
Reported performance of representative VLM
acceleration methods, grouped by their primary
intervention stage.
NExT, MVB, V-MME, ES, LVB, and MLVU denote
NExT-QA, MVBench, Video-MME, EgoSchema,
LongVideoBench, and MLVU, respectively.
Ret./Budget follows the source-defined input,
token, cache, or KV budget and is not directly
comparable across methods.
``--'' denotes an unreported result;
$^\dagger$ identifies a stage-specific component
or configuration of a multi-stage method;
$^{\mathrm{C}}$ denotes a cross-stage method.
Results from different hosts or protocols
should not be interpreted as direct rankings.
}
\label{tab:training_free_benchmark}

\end{table*}

\subsection{Comparison Protocol}
\label{app:comparison}

\noindent\textbf{Evidence Sources.}
Our quantitative analysis combines controlled
experiments with results reported in prior studies.
The controlled experiments evaluate
LLaVA-OneVision-7B with 32 input frames on an
NVIDIA H100 (80 GB) GPU, covering Video-MME,
EgoSchema, LongVideoBench, and MVBench.
Other comparisons draw on original publications
and unified evaluations under shared settings
\cite{lin2026v}.
Derived metrics, including QLB and quality-retention
ratios, are calculated from the corresponding
reported or measured values.

\noindent\textbf{Comparability.}
We prioritize comparisons with matched host models,
benchmarks, input budgets, and measurement protocols.
For our controlled experiments, accuracy and TTFT
are averaged equally across the four benchmarks.
QLB is calculated relative to the matched baseline,
which scores 100. Literature-based comparisons are
interpreted within their respective evaluation groups.
In particular, decoding results using different
task-quality metrics and PTQ results using different
benchmark sets are not directly ranked.
The broader results in Table~\ref{tab:training_free_benchmark}
serve as a reference rather than a unified ranking.

\noindent\textbf{Realized Efficiency.}
We distinguish reductions in frames, visual tokens,
or KV-cache size from measured latency and memory
savings. Our TTFT measurements cover model-side
processing, including visual encoding and prefill,
but exclude external video preprocessing and
subsequent decoding.
For literature-based results, we retain the original
measurement boundaries and do not infer system
performance from compression ratios.
Stage-level and end-to-end measurements are
interpreted separately unless a complete timing
breakdown is available.

\section{Scope and Boundaries of Training-Free Acceleration}
\label{app:training_free_scope}

\noindent\textbf{Training-Free Scope.}
We define training-free acceleration as modifying the
inference procedure of a pretrained VLM without
training additional acceleration-specific parameters.
This includes frame and token selection, merging,
computation reuse, attention sparsification, and
KV-cache management. Pretrained auxiliary models
may be used without further training, although
their inference overhead must be accounted for.
Methods requiring a newly trained selector,
predictor, or draft model fall outside our core scope,
as illustrated by Sparrow \cite{zhang2026sparrow}.

\noindent\textbf{Preparation and Calibration.}
Training-free deployment does not necessarily imply
the absence of data-dependent preparation.
For example, FitPrune derives pruning configurations
from attention statistics without parameter training
\cite{ye2025fitprune}. In contrast, PTQ methods
such as MBQ and QIG use calibration data to determine
quantization parameters or sensitivity
\cite{li2025mbq,xiang2026fine}.
We therefore discuss calibration-dependent PTQ
as a separate cross-stage extension rather than
conflating it with directly applicable
training-free inference methods.

\noindent\textbf{Output Fidelity.}
Training dependency and output fidelity are
independent considerations.
Exact speculative decoding can preserve the
target model's output distribution under its
specified verification and correction rules
\cite{leviathan2023fast,chen2023accelerating}.
In contrast, relaxed verification methods such
as LVSpec improve acceptance by allowing
semantically plausible alternatives, without
guaranteeing identical output distributions
\cite{ji2026see}.
Thus, training-free acceleration does not
necessarily imply lossless inference.

\section{Workload Shifts: Streaming and Multi-Turn Inference}
\label{app:workloads}

Most training-free VLM acceleration studies are evaluated
under offline, single-turn VideoQA settings, where the
complete video is available before answering a single
query. However, streaming and multi-turn workloads
introduce additional constraints on inference efficiency.
Streaming requires causal processing of incoming frames
and bounded historical memory, whereas multi-turn
interaction creates opportunities for computation reuse
while increasing the cost of maintaining dialogue context.
These settings change how visual evidence is accessed,
retained, and reused, making conventional accuracy and
single-query latency insufficient to characterize
deployment efficiency. We examine these two workload
shifts and their implications for training-free
acceleration.

\subsection{Offline-to-Streaming Workload Shift}
\label{app:streaming}

Offline VideoQA assumes access to the complete video
before answering a query, whereas streaming VLMs must
process visual evidence causally as frames arrive.
This shift introduces constraints beyond task quality
and query latency: historical memory must remain
bounded, and frame processing must keep pace with
the incoming stream \cite{tang2026streamingeval}.
A necessary condition for avoiding sustained
processing backlogs is
\begin{equation}
f_{\mathrm{proc}}^{\mathrm{sust}}
\geq f_{\mathrm{in}},
\label{eq:stream_feasibility}
\end{equation}
where $f_{\mathrm{in}}$ denotes the frame rate
ingested by the model after sampling, and
$f_{\mathrm{proc}}^{\mathrm{sust}}$ is its sustainable
processing rate. This condition alone does not
guarantee bounded memory or low response latency.

To illustrate this workload shift, we consider two
streaming benchmarks. StreamingBench evaluates
real-time video understanding under continuously
arriving inputs \cite{lin2024streamingbench},
whereas OVO-Bench emphasizes timestamp-aware
reasoning, including backward tracing, real-time
perception, and forward active responding
\cite{niu2025ovobench}.

Table~\ref{tab:offline_streaming_shift}
presents the offline and streaming results reported
by HERMES \cite{zhang2026hermes}.
Each backbone is evaluated with and without the
training-free adaptation. Since the workloads
differ in input access and evaluation protocol,
we interpret changes within each benchmark rather
than comparing absolute scores across benchmarks.

\begin{table*}[t]
\centering
\footnotesize
\setlength{\tabcolsep}{6.0pt}
\renewcommand{\arraystretch}{1.10}

\begin{tabular}{lccccc}
\toprule
\textbf{Backbone / Setting} &
\multicolumn{3}{c}{\textbf{Offline Video Understanding}} &
\multicolumn{2}{c}{\textbf{Streaming Video Understanding}} \\
\cmidrule(lr){2-4}
\cmidrule(lr){5-6}
&
\textbf{MVBench$\uparrow$} &
\textbf{EgoSchema$\uparrow$} &
\textbf{V-MME Avg.$\uparrow$} &
\textbf{SBench$\uparrow$} &
\textbf{OVO Avg.$\uparrow$} \\
\midrule

\rowcolor{gray!10}
\multicolumn{6}{c}{
\textit{\textbf{LLaVA-OneVision-7B}}} \\

Base
& 57.02 & 59.93 & 57.67 & 71.34 & 53.35 \\

+ HERMES (4K)
& 56.92 & 60.29 & 58.85 & 73.23 & 58.27 \\

\midrule

\rowcolor{gray!10}
\multicolumn{6}{c}{
\textit{\textbf{Qwen2.5-VL-7B}}} \\

Base
& 65.00 & 58.47 & 64.52 & 73.31 & 52.28 \\

+ HERMES (4K)
& 65.53 & 59.97 & 60.63 & 79.44 & 59.21 \\

\bottomrule
\end{tabular}

\caption{
Offline and streaming video understanding
performance reported by HERMES (\%).
SBench denotes the real-time visual understanding
evaluation of StreamingBench.
OVO Avg. averages the real-time visual perception
and backward tracing categories of OVO-Bench.
HERMES (4K) specifies a 4K-token KV-cache budget.
Offline and streaming benchmarks use different
evaluation protocols; scores are compared only
within the same benchmark.
}
\label{tab:offline_streaming_shift}
\end{table*}

The results reveal a workload-dependent trade-off.
On LLaVA-OneVision-7B, HERMES improves OVO-Bench
from 53.35 to 58.27, while offline scores change
only modestly. On Qwen2.5-VL-7B, StreamingBench
and OVO-Bench improve by 6.13 and 6.93 points,
respectively, but Video-MME decreases by 3.89
points. Thus, adaptation to causal streaming
may improve online understanding without
uniformly preserving offline performance.

SimpleStream further shows that a short recent-frame
window can form a strong training-free streaming
baseline \cite{shen2026simplestream}.
Other methods target complementary costs:
StreamingTOM compresses streaming visual context
\cite{chen2026streamingtom}, whereas STC reduces
redundant visual encoding through feature reuse
\cite{wang2026accelerating}.
Together, these studies suggest that streaming
acceleration should be assessed through causal
task performance, memory growth, processing
throughput, and response latency, rather than
offline accuracy alone.

\subsection{Single-Turn to Multi-Turn Interaction}
\label{app:multiturn}

When multiple questions refer to the same video,
inference becomes a stateful interaction rather
than a sequence of independent requests.
Previously computed visual features may be reused,
but dialogue history introduces additional
context and memory costs. Consequently,
training-free multi-turn acceleration must balance
computation reuse with the preservation of
relevant visual and textual evidence.

For a fixed video with one-time preparation cost
$\tau_{\mathrm{prep}}$ and $Q$ successive queries,
the amortized processing time per query can be
expressed as
\begin{equation}
\bar{\tau}_{Q}
=
\frac{
\tau_{\mathrm{prep}}
+
\sum_{q=1}^{Q}
\left(
\tau_{\mathrm{adapt}}^{(q)}
+
\tau_{\mathrm{infer}}^{(q)}
\right)
}{Q},
\label{eq:amortized_query}
\end{equation}
where $\tau_{\mathrm{adapt}}^{(q)}$ denotes the
query-specific cost of selection, retrieval,
compression, or state reconstruction, and
$\tau_{\mathrm{infer}}^{(q)}$ denotes the remaining
inference time. This expression characterizes
the potential benefit of reuse; it is not a
measured latency result in this survey.

Training-free methods exploit this opportunity
at different stages. STTM performs query-agnostic
spatiotemporal merging, potentially allowing
compressed visual features to serve multiple
queries \cite{hyun2025multi}. ReKV instead
reuses previously computed KV states and
retrieves query-relevant historical information
\cite{di2025streaming}. However, aggressive
reuse or eviction can also discard evidence
needed by subsequent questions.

SVBench evaluates video understanding under
single-instance and temporal multi-turn dialogue
settings \cite{yang2025svbench}.
Its evaluation includes METEOR, a reference-based
text generation metric \cite{banerjee2005meteor},
and GPT4-Score, which uses GPT-4 to assess
the semantic accuracy of generated answers
against reference responses
\cite{yang2025svbench}.
To describe the observed difference between
the two evaluation settings, we calculate
\begin{equation}
\Delta_{\mathrm{ctx}}
=
S_{\mathrm{multi}}
-
S_{\mathrm{single}},
\label{eq:history_gain}
\end{equation}
where $S_{\mathrm{multi}}$ and
$S_{\mathrm{single}}$ are the corresponding
benchmark scores.
Table~\ref{tab:dialogue_gain} presents the
reported scores and our derived differences.

\begin{table*}[t]
\centering
\footnotesize
\setlength{\tabcolsep}{4.8pt}
\renewcommand{\arraystretch}{1.10}

\begin{tabular}{lrrrrrr}
\toprule
\textbf{Model} &
\multicolumn{3}{c}{\textbf{METEOR}} &
\multicolumn{3}{c}{\textbf{GPT4-Score}} \\
\cmidrule(lr){2-4}
\cmidrule(lr){5-7}
&
\textbf{Single$\uparrow$} &
\textbf{Multi$\uparrow$} &
\textbf{$\Delta_{\mathrm{ctx}}$} &
\textbf{Single$\uparrow$} &
\textbf{Multi$\uparrow$} &
\textbf{$\Delta_{\mathrm{ctx}}$} \\
\midrule

InternVL2
& 26.85 & 32.74 & +5.89
& 32.26 & 42.25 & +9.99 \\

VILA
& 28.44 & 33.57 & +5.13
& 34.63 & 45.41 & +10.78 \\

InternLM-XComposer2.5
& 23.32 & 31.02 & +7.70
& 37.11 & 49.69 & +12.58 \\

MiniCPM-V 2.6
& 27.36 & 33.90 & +6.54
& 36.73 & 48.57 & +11.84 \\

GPT-4V
& 28.66 & 36.47 & +7.81
& 44.94 & 60.11 & +15.17 \\

GPT-4o
& 28.81 & 36.84 & +8.03
& 45.37 & 60.70 & +15.33 \\

\bottomrule
\end{tabular}

\caption{
Single-instance (Single) and multi-turn (Multi)
video QA performance on SVBench.
METEOR measures reference-based text similarity,
including synonym and stemming matches;
GPT4-Score evaluates answer correctness
using GPT-4 as a judge.
Both metrics are shown on the reported
0--100 score scale.
$\Delta_{\mathrm{ctx}}$ is the derived difference
between Multi and Single, in score points.
It does not measure inference acceleration
or isolate the causal effect of dialogue history.
}
\label{tab:dialogue_gain}
\end{table*}

All listed models achieve higher METEOR and
GPT4-Score values under multi-turn evaluation.
For GPT-4o, METEOR increases from 28.81 to
36.84 and GPT4-Score from 45.37 to 60.70.
These results suggest that preceding dialogue
context can provide useful information for
subsequent video understanding.
However, SVBench evaluates task quality rather
than the computation or memory costs of
maintaining that context.

The implication for training-free acceleration
is therefore not to retain all historical states,
but to selectively preserve useful evidence
while reducing repeated computation.
For repeated queries over a fixed video,
amortized latency and persistent memory
are important complements to per-query
accuracy and latency.

Overall, streaming and multi-turn interaction
introduce distinct workload constraints.
Streaming requires causal processing and
sustainable state management, whereas
multi-turn interaction creates opportunities
for reuse alongside growing dialogue history.
Both settings reinforce the need to evaluate
training-free acceleration under its intended
workload, with task quality and realized
system costs considered together.

\section{Speculative Decoding for VLM Acceleration}
\label{app:speculative}

Speculative decoding (SD) accelerates autoregressive
generation by allowing a draft model to propose
multiple tokens for parallel verification by a
target model. For VLMs, however, drafting can be
expensive because both models must process long
visual contexts. Effective acceleration therefore
depends on draft cost, draft--target alignment,
verification overhead, and accepted progress
\cite{zhang2025sparse,ji2025specvlm}.

\subsection{Draft--Verify Efficiency}
\label{app:spec_efficiency}

Training-free VLM SD follows several complementary
directions. Sparse-to-Dense (StD) combines sparse
Top-$K$ drafting with dense verification, reporting
up to $1.94\times$ wall-clock speedup without
changing the target distribution
\cite{zhang2025sparse}.
SpecVLM instead reduces draft-side visual tokens
through verifier-guided pruning while retaining
full-context target verification. It removes up
to 90\% of draft visual tokens and reports up to
$2.68\times$ and $2.11\times$ decoding speedup on
LLaVA-OneVision-72B and Qwen2.5-VL-32B,
respectively \cite{ji2025specvlm}.

ParallelVLM further overlaps drafting and
verification, using Unbiased Verifier-Guided
Pruning to improve draft--target alignment.
It reports average speedups of $3.36\times$
and $2.42\times$ on the same target models
\cite{kong2026parallelvlm}.
LVSpec takes a different approach by relaxing
verification for less visually relevant tokens
while strictly checking visual-semantic anchors.
Its position-shift-tolerant mechanism increases
accepted progress without guaranteeing exact
output-distribution preservation
\cite{ji2026see}.

These methods expose three acceleration levers:
reducing draft computation, overlapping execution,
and increasing accepted progress. Their combined
effect must be assessed through measured decoding
efficiency rather than acceptance alone.

\subsection{Empirical Trade-offs}
\label{app:spec_tradeoff}

Cross-paper speedups cannot be directly compared
because draft--target pairs, hardware, speculative
windows, and workloads differ.
We therefore organize matched comparisons reported
by ParallelVLM and LVSpec
\cite{kong2026parallelvlm,ji2026see}.
The former evaluates five video benchmarks,
including Video Detail Caption
\cite{chai2025auroracap}, Video-MME, MVBench,
MLVU, and LongVideoBench
\cite{fu2025video,li2024mvbench,
zhou2025mlvu,wu2024longvideobench}.
The latter reports results on Video Detail Caption.
Table~\ref{tab:spec_matched} summarizes the
corresponding source-reported measurements.

Let $M$ denote mean accepted length and $S$
the decoding speedup relative to autoregressive
generation. We calculate the relative improvements
over the SpecVLM baseline within each matched
block as
\begin{equation}
\Delta M
=
\frac{M_{\mathrm{new}}-M_{\mathrm{base}}}
{M_{\mathrm{base}}},
\qquad
\Delta S
=
\frac{S_{\mathrm{new}}-S_{\mathrm{base}}}
{S_{\mathrm{base}}}.
\label{eq:spec_relative_gain}
\end{equation}

\begin{table*}[t]
\centering
\footnotesize
\setlength{\tabcolsep}{4.8pt}
\renewcommand{\arraystretch}{1.10}

\begin{tabular}{llccccc}
\toprule
\textbf{Target / Draft Setting} &
\textbf{Method} &
\textbf{$M\uparrow$} &
\textbf{Speedup$\uparrow$} &
\textbf{$\Delta M\uparrow$} &
\textbf{$\Delta S\uparrow$} &
\textbf{Fidelity} \\
\midrule

\rowcolor{gray!10}
\multicolumn{7}{c}{
\textit{\textbf{Five-benchmark matched evaluation in ParallelVLM}}
} \\

LLaVA-OV 72B / 7B
& SpecVLM
& 4.42
& 2.74$\times$
& --
& --
& Exact \\

LLaVA-OV 72B / 7B
& ParallelVLM
& \textbf{6.83}
& \textbf{3.36$\times$}
& \textbf{+54.4\%}
& \textbf{+22.6\%}
& Exact \\

\addlinespace[2pt]

Qwen2.5-VL 32B / 7B
& SpecVLM
& 4.19
& 2.11$\times$
& --
& --
& Exact \\

Qwen2.5-VL 32B / 7B
& ParallelVLM
& \textbf{4.28}
& \textbf{2.42$\times$}
& \textbf{+2.2\%}
& \textbf{+14.7\%}
& Exact \\

\midrule

\rowcolor{gray!10}
\multicolumn{7}{c}{
\textit{\textbf{Video Detail Caption evaluation in LVSpec}}
} \\

Qwen2.5-VL 32B / 7B
& SpecVLM
& 3.29
& 2.00$\times$
& --
& --
& Exact \\

Qwen2.5-VL 32B / 7B
& LVSpec
& \textbf{7.76}
& \textbf{2.70$\times$}
& \textbf{+135.9\%}
& \textbf{+35.0\%}
& 99.6\% \\

\addlinespace[2pt]

LLaVA-OV 72B / 7B
& SpecVLM
& 3.37
& 2.38$\times$
& --
& --
& Exact \\

LLaVA-OV 72B / 7B
& LVSpec
& \textbf{7.34}
& \textbf{2.94$\times$}
& \textbf{+117.8\%}
& \textbf{+23.5\%}
& 99.8\% \\

\bottomrule
\end{tabular}

\caption{
Matched-protocol speculative decoding results.
Target/Draft identifies the verifier and drafter
model sizes, respectively.
$M$ is the mean number of tokens accepted per
speculative iteration.
Speedup is the source-reported decoding acceleration
relative to autoregressive generation ($1.00\times$).
$\Delta M$ and $\Delta S$ are percentage improvements
over SpecVLM within each matched block;
the five-benchmark $\Delta M$ values are calculated
from unrounded averages.
The first block covers Video Detail Caption,
Video-MME, MVBench, MLVU, and LongVideoBench;
the second uses Video Detail Caption (VDC).
``Exact'' indicates strict target-distribution
preservation, whereas percentages indicate
task-performance retention under relaxed verification,
not exact output agreement.
Results are comparable only within matched blocks.
}
\label{tab:spec_matched}
\end{table*}

Table~\ref{tab:spec_matched} shows that accepted
progress and decoding speedup are not proportional.
For LLaVA-OneVision-72B/7B, ParallelVLM improves
mean accepted length by 54.4\% but speedup by
22.6\%. On Qwen2.5-VL-32B/7B, accepted length
increases by only 2.2\%, while speedup improves
by 14.7\%. This difference reflects the additional
contribution of overlapping execution
\cite{kong2026parallelvlm}.

LVSpec illustrates the fidelity--efficiency
trade-off of relaxed verification. On
Qwen2.5-VL-32B/7B, it increases accepted
length from 3.29 to 7.76 and speedup from
$2.00\times$ to $2.70\times$, with 99.6\%
task-performance retention. On
LLaVA-OneVision-72B/7B, the corresponding
speedup rises from $2.38\times$ to
$2.94\times$, with 99.8\% retention
\cite{ji2026see}.
Unlike strict SD, these gains do not imply
preservation of the original output distribution.

\subsection{Evaluation and Net Gain}
\label{app:spec_evaluation}

Speculation does not necessarily yield acceleration.
In the self-speculative settings reported by LVSpec,
naive SD achieves only $0.88\times$, $0.94\times$,
and $0.86\times$ speedup on Video Detail Caption,
Video Detail Description, and MovieChat with
Qwen2.5-VL. A LLaVA-OneVision configuration
falls to $0.65\times$ on MovieChat
\cite{ji2026see,song2024moviechat}.
These results show that drafting and verification
overhead can exceed the computation saved.

A useful diagnostic is the wall-clock cost per
committed output token:
\begin{equation}
C_{\mathrm{prog}}
=
\frac{
\mathbb{E}[T_{\mathrm{iter}}^{\mathrm{wall}}]
}{
\mathbb{E}[N_{\mathrm{commit}}]
},
\label{eq:spec_progress_cost}
\end{equation}
where $T_{\mathrm{iter}}^{\mathrm{wall}}$
includes drafting, verification, synchronization,
and control overhead, accounting for any execution
overlap. $N_{\mathrm{commit}}$ is the number of
output tokens committed per iteration.
This formulation avoids incorrectly adding
draft and verification times when they execute
concurrently.

The effect on request-level latency also depends
on output length. Under a simplified timing model
that excludes additional system overhead, the
decoding-time fraction can be approximated as
\begin{equation}
\alpha_{\mathrm{dec}}^{\mathrm{proxy}}(N_o)
=
\frac{
(N_o-1)\tau_{\mathrm{TPOT}}
}{
\tau_{\mathrm{TTFT}}
+
(N_o-1)\tau_{\mathrm{TPOT}}
},
\label{eq:spec_decode_fraction}
\end{equation}
where $N_o$ is the generated-token count.
For fixed TTFT and TPOT, longer responses
increase the share of time available for
decoding-side optimization. However, this
quantity is not a measured E2E latency fraction:
video preprocessing, scheduling, and other
system overhead may remain outside its scope.

Consequently, SD evaluations should report
draft--target configurations, visual budgets,
speculative windows, mean accepted length,
decoding throughput or TPOT, output length,
and verification fidelity. Where possible,
drafting and verification costs, TTFT, and
request-level E2E latency should also be measured
\cite{zhang2025sparse,ji2025specvlm,
kong2026parallelvlm,ji2026see}.
The objective is not maximum acceptance alone,
but reliable conversion of speculative progress
into realized inference efficiency.

\end{document}